\documentclass[lettersize,journal]{IEEEtran}
\usepackage{amsmath,amsfonts}
\usepackage{algorithmic}
\usepackage{algorithm}
\usepackage{array}
\usepackage[caption=false,font=normalsize,labelfont=sf,textfont=sf]{subfig}
\usepackage{textcomp}
\usepackage{stfloats}
\usepackage{url}
\usepackage{verbatim}
\usepackage{graphicx}
\usepackage{cite}
\usepackage{booktabs}
\usepackage{float}
\usepackage{orcidlink}

\begin{document}

\title{Shot Sequence Ordering for Video Editing: Benchmarks, Metrics, and Cinematology-Inspired Computing Methods}

\author{
  Yuzhi Li\orcidlink{0009-0009-7050-7775}, Haojun Xu\orcidlink{0000-0001-8312-6634}, Feng Tian\orcidlink{0009-0003-9904-6729}

\thanks{\textit{Corresponding Author: Feng Tian}}
\thanks{Yuzhi Li, Haojun Xu and Feng Tian are with Shanghai University, Shanghai, China, 200072, (e-mails: shadowmcv@shu.edu.cn; sacross\_k@shu.edu.cn; ouman@shu.edu.cn)}

\thanks{This work has been submitted to the IEEE for possible publication. Copyright may be transferred without notice, after which this version may no longer be accessible.}
}

\markboth{Journal of \LaTeX\ Class Files,~Vol.~14, No.~8, August~2021}%
{Shell \MakeLowercase{\textit{et al.}}: A Sample Article Using IEEEtran.cls for IEEE Journals}

\IEEEpubid{}

\maketitle

\begin{abstract}
With the rising popularity of short video platforms, the demand for video production has increased substantially. However, high-quality video creation continues to rely heavily on professional editing skills and a nuanced understanding of visual language. To address this challenge, the Shot Sequence Ordering (SSO) task in AI-assisted video editing has emerged as a pivotal approach for enhancing video storytelling and the overall viewing experience. Nevertheless, the progress in this field has been impeded by a lack of publicly available benchmark datasets. In response, this paper introduces two novel benchmark datasets, AVE-Order and ActivityNet-Order. Additionally, we employ the Kendall Tau distance as an evaluation metric for the SSO task and propose the Kendall Tau Distance-Cross Entropy Loss. We further introduce the concept of Cinematology Embedding, which incorporates movie metadata and shot labels as prior knowledge into the SSO model, and constructs the AVE-Meta dataset to validate the method's effectiveness. Experimental results indicate that the proposed loss function and method substantially enhance SSO task accuracy. All datasets are publicly accessible at our github repository\footnote{https://github.com/litchiar/ShotSeqBench}.
\end{abstract}


\section{Introduction}
\begin{figure}[ht]
 \centering 
 \includegraphics[width=1\columnwidth]{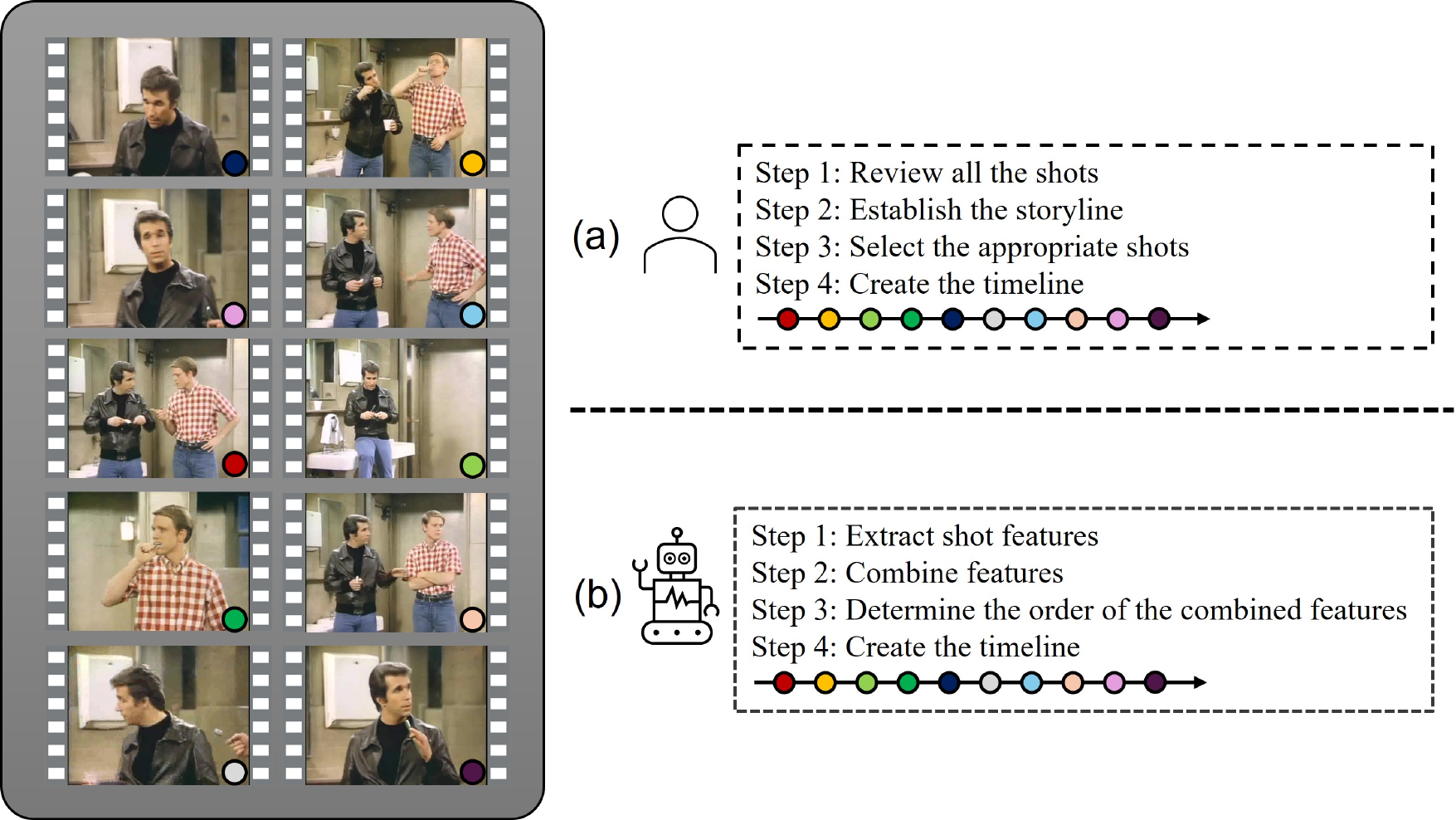}
 \caption{(a) Workflow of professional video editors in traditional editing, involving manual arrangement of unordered shot clips to construct a coherent, narrative-rich video. (b) Application of AI-assisted systems in video editing, where AI analyzes shot content and intelligently arranges them, enabling users to generate a coherent shot sequence efficiently.}
 \label{fig:1}
\end{figure}

With the rapid rise of video media, particularly short videos, an increasing number of users are engaging in video production \cite{cite1}. However, producing high-quality videos requires not only professional knowledge and skills but also a deep understanding of visual language, narrative structure, and audience psychology \cite{cite2}. As depicted in Figure \ref{fig:1}(a), professional video editors often combine numerous unordered shot clips on a timeline to create a coherent and expressive work, a process demanding both technical proficiency and a strong sense of artistic perception.


To reduce the barriers to video editing and improve efficiency, artificial intelligence (AI) technology has been increasingly introduced into various aspects of video production. In recent years, AI-assisted video editing has gained considerable attention \cite{cite5}. As shown in Figure \ref{fig:1}(b), automated video editing tools can help users quickly organize and edit shot clips on the timeline.

Among the various AI video editing tasks, Shot Sequence Ordering (SSO) \cite{cite6} is particularly important, as it directly influences a video's narrative flow and overall visual appeal. The core goal of the SSO task is to automatically arrange video clips in a logical sequence, thereby enhancing the video’s coherence and viewing experience. Consequently, improving the accuracy and intelligence of SSO is vital for advancing AI-assisted video editing.

However, progress in SSO research has been hindered by the lack of publicly available benchmark datasets. An effective shot sequence dataset should contain complete video files and corresponding shot segmentation to enable researchers to analyze temporal relationships and narrative logic accurately. Yet, due to copyright restrictions, many existing long-video datasets (e.g., HowTo100M \cite{cite7} and YouTube-8M \cite{cite8}) only provide feature files or video frames, limiting further exploration and innovation in the SSO task.

To address this issue, we processed existing video datasets to construct two publicly available benchmark datasets: AVE-Order and ActivityNet-Order. The AVE-Order dataset, organized from the AVE dataset \cite{cite6}, has been widely used in various video understanding tasks \cite{cite9,cite10} but lacked labels suitable for SSO. We annotated the original data to create labels specific to the SSO task. Similarly, the ActivityNet dataset \cite{cite11} was used to construct ActivityNet-Order by segmenting complete video files, removing invalid segments, and generating corresponding shot sequence data.

Previous studies \cite{cite6,cite12} treated SSO as a conventional video classification task. However, we found that SSO labels possess a unique structure and hierarchy, rather than being mere classification labels. For instance, in a typical 6-class video task, the system must classify the video into one of several categories (e.g., football, basketball, volleyball, etc.). If a "football" video is misclassified as "basketball" or "badminton," it’s unclear which error is closer to the correct label. In contrast, the SSO task requires more nuanced ordering; for example, if a video sequence labeled "123" is misclassified as "231," this error is closer to the correct order than "321" because it captures the partial sequence relationship ("23"). This distinction indicates that SSO tasks inherently involve hierarchical and relative quality, beyond simple classification.

Based on these observations, we introduced the Kendall Tau distance as an evaluation metric for the SSO task, which effectively measures ordering differences by calculating inconsistencies between shot pairs, quantifying the gap between the model's predicted and true orderings. Building on this, we proposed a new loss function, the Kendall Tau Distance-Cross Entropy (KTD-CE) Loss, which preserves probability distribution information from classification tasks while optimizing local sequence ordering relationships more precisely.

Another key challenge in SSO is ensuring professionalism. While user-generated content (UGC) videos make it difficult to determine professional shot ordering standards, professionally edited films inherently demonstrate high-quality shot ordering. Previous studies \cite{cite13,cite14} have revealed a strong relationship between film genres and shot types, where different genres follow specific shot languages, and shot types are closely tied to narrative structures. Hence, we introduced the concept of Cinematology Embedding, embedding movie metadata and shot types as prior knowledge into the SSO model, to enhance the shot sequence ordering task.

For the movie metadata, we constructed the AVE-Meta dataset, which we collected from IMDb and linked to the AVE dataset \cite{cite6}, incorporating key information such as film genre, director, and release year. Since manually labeling all shots is labor-intensive, we leveraged our previous research \cite{cite15} to generate shot labels using automated methods. We then applied a Video Transformer-based architecture to transform the movie metadata and shot category information into additional input tokens. Experimental results confirmed that this approach significantly improved the accuracy of the task. Overall, the findings demonstrate that the proposed KTD-CE Loss and Cinematology Embedding were highly effective in enhancing SSO.

The main contributions of this study are as follows:
\begin{itemize}
    \item We constructed two benchmark datasets for the SSO task—AVE-Order and ActivityNet-Order—addressing the scarcity of datasets in the SSO field.
    \item We introduced Kendall Tau Distance as a metric for evaluating the accuracy of SSO, effectively capturing the differences between the model's predicted sequence and the true sequence.
    \item We developed the Kendall Tau Distance-Cross Entropy Loss, specifically tailored for the SSO task, which significantly enhances model performance by integrating ordering relationships with classification information.
    \item We created the AVE-Meta dataset, linking movie metadata with movie scenes, thus providing a new foundation for movie analysis tasks.
    \item We proposed Cinematology Embedding, which incorporates movie metadata and shot labels as prior knowledge into the SSO model, thereby improving shot sequence ordering accuracy.
\end{itemize}

The structure of this paper is as follows: Section \ref{sec:2} reviews related work; Section \ref{sec:3} details the two shot sequence ordering datasets; Section \ref{sec:4} introduces the Kendall Tau-Cross Entropy Loss; Section \ref{sec:5} presents the Cinematology Embedding method; Section \ref{sec:6} outlines the experimental details and results; Section \ref{sec:7} discusses challenges and future research directions for the SSO task; and Section \ref{sec:8} concludes the paper.

\section{Related Work} \label{sec:2}

\subsection{AI-Assisted Video Editing} \label{sec:2.1}

AI-assisted video editing can be broadly categorized into two main areas: visual editing and timeline editing. Visual editing focuses on modifying and enhancing video content, such as lighting enhancement \cite{cite16}, video inpainting \cite{cite17}, object removal \cite{cite18}, and style transfer \cite{cite20,cite21}. Generative models, particularly diffusion models \cite{cite22} and visual large language models (vLLMs) \cite{cite25,cite26}, are widely employed in this field, offering substantial performance improvements over traditional methods. Prompt-based editing methods, in particular, have enhanced the flexibility and accuracy of visual editing.

Timeline editing primarily addresses the organization and adjustment of videos in the temporal dimension, including tasks like automatic video summarization \cite{cite28}, trailer generation \cite{cite29}, and rearranging clips based on specific requirements \cite{cite30}. Current research predominantly targets specific applications, such as generating dialogue videos from scripts \cite{cite31} or creating trailers from movie content \cite{cite32, cite33}. However, there remains a lack of comprehensive research on more general video editing needs, especially for non-professional users (e.g., personal or short videos). This study serves as an initial exploration of general timeline editing.

\subsection{Shot Sequence Ordering} \label{sec:2.2}
In video production, shot order is critical for storytelling and emotional impact. A well-organized shot sequence not only engages the audience but also enhances the clarity of the video content \cite{cite34}. As an inherent feature of videos, shot order has been employed in numerous self-supervised learning tasks. For instance, Xiao et al. used randomly shuffled shot sequences as supervision signals to train models in learning effective spatiotemporal visual representations \cite{cite35}. Similarly, Kim et al. introduced a self-supervised learning method requiring models to restore shuffled shot orders to capture spatiotemporal features in videos \cite{cite36}. Additionally, Xu et al. extracted valuable spatiotemporal representations from large-scale unlabeled videos by predicting shuffled shot orders \cite{cite37}.

The goal of SSO is to arrange unordered shots into a temporal sequence that effectively conveys a coherent storyline or visual flow \cite{cite6}. As AI-generated content (AIGC) continues to evolve, video generation techniques \cite{cite38,cite40} have emerged as a prominent research area in computer vision. In this process, numerous candidate shots are generated and must be meticulously edited and arranged to form a coherent narrative. Shot sequence ordering helps select the most logical and aesthetically pleasing shot arrangements from various options. To our knowledge, this study is the first to systematically investigate the SSO task.
\subsection{Learning Representations from Movies} \label{sec:2.3}
Compared to ordinary videos, movies, as high-quality video resources, typically exceed one hour in duration, making visual model learning from movie data particularly challenging. Due to copyright restrictions, obtaining complete movie content for training and experimentation is often difficult. Despite this, research on movie data has led to the development of several movie datasets for different visual tasks. Examples include Hollywood2 for action recognition \cite{cite41}, OVSD for scene segmentation \cite{cite51}, MovieShots, CineScale \cite{cite42}, Historian \cite{cite43}, and HistShot \cite{cite44} for shot analysis, MAD for audio description \cite{cite46}, LSMDC for movie narration description \cite{cite47}, and MovieQA for question answering \cite{cite48}.

Additionally, some large-scale movie datasets support multiple visual tasks simultaneously. The first large-scale movie dataset, MovieNet, offers role labels, action and scene labels, shot labels, and scripts from 1,100 movies \cite{cite49}. The LVU dataset contains over 1,000 hours of video and covers nine distinct tasks \cite{cite50}, while the AVE dataset includes 5,500 movie scene videos collected from the MovieCLIP channel, supporting five different visual tasks \cite{cite6}.

In this study, we have compiled two Shot Sequence Ordering datasets that provide complete shot and sequence labels. We also introduced the AVE-Meta dataset, which is closely related to AVE and includes additional movie metadata labels. Building on this, we propose the Cinematology Embedding method, which enhances model learning of potential shot grammar by leveraging shot analysis.

\section{SSO Datasets} \label{sec:3}

\subsection{Problem definition} \label{sec:3.1}
To model shot sequence ordering, we define it as a video classification task. Given an unordered set of shots $S=\{s_1,s_2,\ldots,s_k\}$, the objective is to predict the sequence label for this set. For $k$ shots, there are $k!$ (factorial of $k$) possible permutations. Following the method in \cite{cite6, cite35}, we set $k=3$ , treating the task as a 6-class classification problem, with each class representing one of the possible orderings of the three shots.

This study focuses on 3-shot sequence ordering. Future research will explore cases where $k>3$, as the factorial growth in permutations with increasing $k$ significantly raises task complexity.

\subsection{AVE-Order} \label{sec:3.2}

The AVE-Order dataset was constructed based on the AVE dataset with additional organization. Although \cite{cite6} offered an initial exploration of shot sequence ordering, it lacked specific shot sequences and corresponding label information. We divided the dataset into training, validation, and test sets in a 7:1:2 ratio, ensuring that the scenes in each set were non-overlapping. Shot sequences were randomly shuffled following a uniform distribution, and corresponding sorting labels were generated. Table \ref{tab:1} provides the statistical details of AVE-Order, while Figure \ref{fig:2} (left) illustrates its label distribution.

\begin{table}[h]
\centering
\caption{Statistics of AVE-Order}
\label{tab:1}
\begin{tabular}{lllll}
\toprule
                             & Train  & Val    & Test   & Total  \\
                             \midrule
Num. of scenes               & 3876   & 555    & 1106   & 5537   \\
Num. of   sequences          & 44329  & 6135   & 12721  & 63185  \\
Avg. duration of   scenes    & 133.64 & 134.63 & 134.08 & 133.83 \\
Avg. duration of   sequences & 10.95  & 11.34  & 10.93  & 10.99 \\
\bottomrule
\end{tabular}
\end{table}

Notably, although label information for shot sequences was available in the training set, we applied data augmentation by randomly shuffling sequences and generating new labels during training. The test set labels were retained to ensure that the evaluation metrics could objectively assess the SSO model's performance.

\subsection{ActivityNet-Order} \label{sec:3.3}

After reviewing existing video analysis datasets, we selected ActivityNet 100 \cite{cite11} as the foundation for the shot sequence ordering dataset. The videos in this dataset are sourced from high-quality, user-uploaded content on YouTube, with most videos comprising multiple complete shots. We used the Transnet v2 \cite{cite53} model to segment the videos into distinct shots, followed by further cleaning to remove invalid shots, such as mis-segmented clips or transitional black screens. The shot sequences were then randomly shuffled using a uniform distribution, and corresponding sorting labels were generated, resulting in 14,490 shot sequence samples from approximately 4,000 scenes. The shot segmentation data is available in the project's code repository.

\begin{figure}[h]
 \centering 
 \includegraphics[width=1\columnwidth]{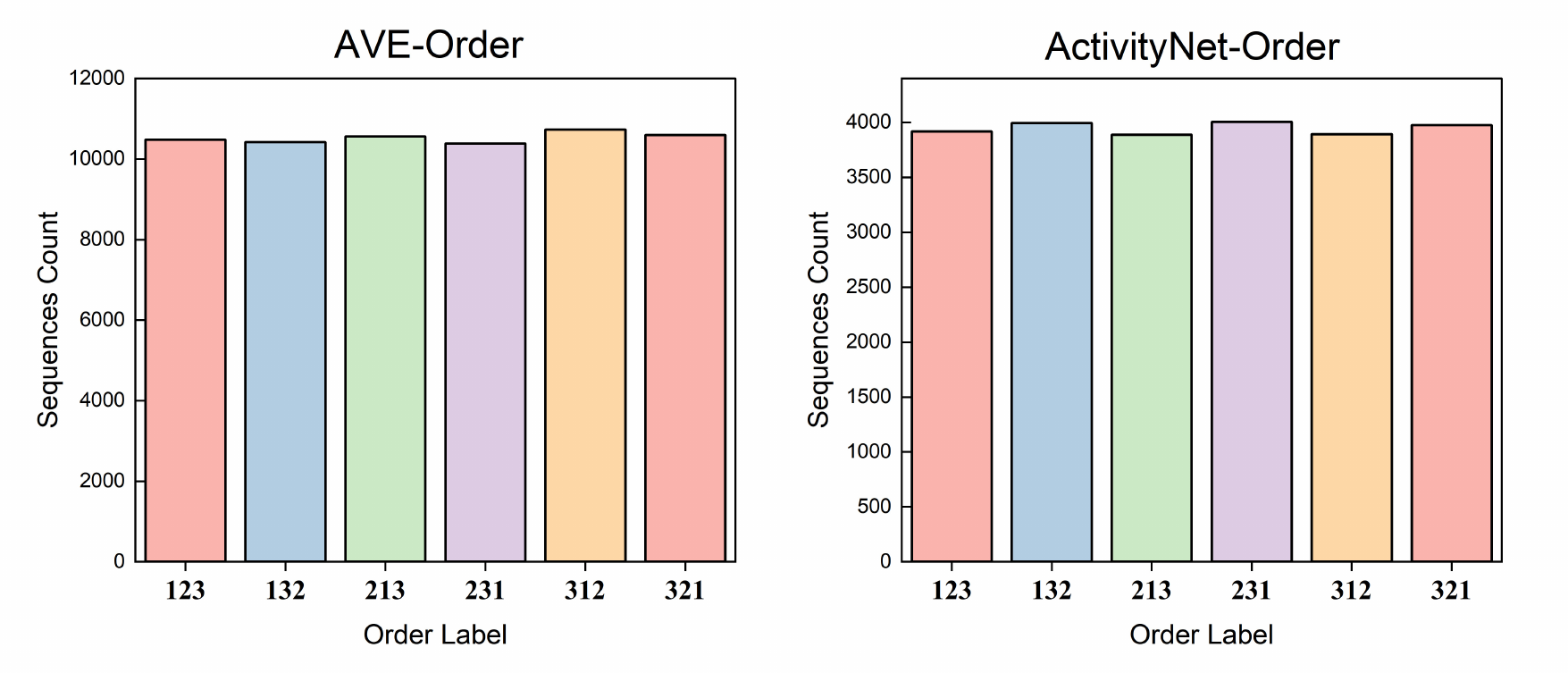}
 \caption{The label distribution of AVE-Order and ActivityNet-Order.}
 \label{fig:2}
\end{figure}

We adopted the data split defined in \cite{cite11} to divide the dataset into training, validation, and test sets. Table \ref{tab:2} provides the statistical details of ActivityNet-Order, while Figure \ref{fig:2} (right) displays its label distribution.

\begin{table}[h]
\centering
\caption{Statistics of ActivityNet-Order}
\label{tab:2}
\begin{tabular}{lllll}
\toprule
                             & Train  & Val    & Test   & Total  \\
                             \midrule
Num. of scenes               & 2495   & 398    & 1159   & 4052   \\
Num. of   sequences          & 14490  & 2550   & 6633   & 23673  \\
Avg. duration of   scenes    & 134.68 & 132.89 & 131.34 & 133.55 \\
Avg. duration of   sequences & 21.08  & 18.90  & 20.94  & 20.80 \\
\bottomrule
\end{tabular}
\end{table}

\section{Kendall Tau Distance-Cross Entropy Loss} \label{sec:4}
\subsection{Preliminary} \label{sec:4.1}

The Kendall Tau distance is a statistical measure used to assess the difference between two rankings, commonly employed to compare ranking similarities. It measures this difference by counting the number of "inverted pairs" between the two rankings. Specifically, an "inverted pair" occurs when the order of a pair of elements in one ranking is the reverse of their order in the other.

Given two rankings $\sigma_1$ and $\sigma_2$, each containing $n$  element, the goal of the Kendall Tau distance calculation is to identify the number of element pairs with opposite orders in  $\sigma_1$ and  $\sigma_2$. The calculation steps are as follows:

\begin{algorithm}[H]
\caption{Kendall Tau Distance}
\begin{algorithmic}[1]
\REQUIRE Two rankings $\sigma_1$ and $\sigma_2$, each containing $n$ elements
\ENSURE The Kendall tau distance between $\sigma_1$ and $\sigma_2$
\STATE Let $n$ be the number of elements in $\sigma_1$
\STATE Initialize \textit{count\_inversions} to 0
\FOR{$i = 1$ to $n-1$}
    \FOR{$j = i+1$ to $n$}
        \IF {($\sigma_1[i] < \sigma_1[j]$ and $\sigma_2[i] > \sigma_2[j]$) or ($\sigma_1[i] > \sigma_1[j]$ and $\sigma_2[i] < \sigma_2[j]$)}
            \STATE Increment \textit{count\_inversions} by 1
        \ENDIF
    \ENDFOR
\ENDFOR
\RETURN \textit{count\_inversions}
\end{algorithmic}
\end{algorithm}

$count\_inversions$ is used to calculate the number of inverted pairs, which represents the Kendall Tau distance. If the two rankings are entirely identical, the Kendall Tau distance is 0. In contrast, when the rankings are completely reversed, the Kendall Tau distance reaches its maximum value of $n\left(n-1\right)/2$

\begin{figure}[h]
 \centering 
 \includegraphics[width=0.5\columnwidth]{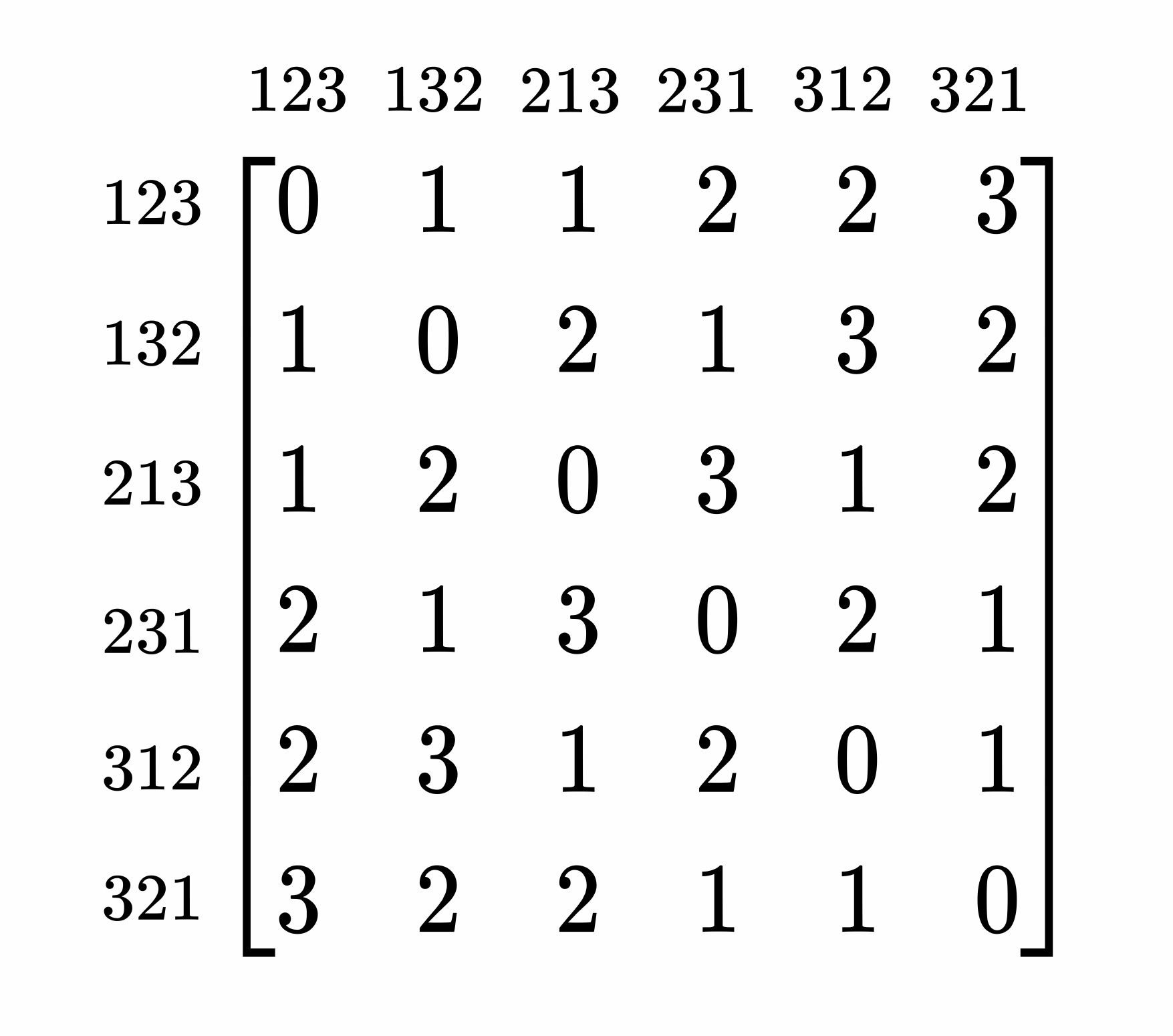}
 \caption{ Kendall Tau distance matrix when $k=3$}
 \label{fig:3}
\end{figure}

To apply the Kendall Tau distance in the SSO task, we represent the various ordering labels as a Kendall Tau distance matrix. Figure \ref{fig:3} displays this matrix for the case where $k=3$ , with each element representing the Kendall Tau distance between two different orderings.

\subsection{KTD-CE Loss} \label{sec:4.2}

SSO is often treated as a video classification task \cite{cite6,cite35}. However, treating each possible order as an independent classification label overlooks the distance relationships between shot ordering labels. For instance, if a model misclassifies the order "123" as "213," this error is less severe than misclassifying it as "312." Thus, relying solely on traditional classification loss may not effectively capture these differences.

To address this, we propose the Kendall Tau Distance-Cross Entropy Loss (KTD-CE Loss), a hybrid loss function that combines cross-entropy loss with Kendall Tau distance loss. This approach accounts for both the discrete classification information of labels and the inherent distance relationships in shot ordering. The torch-style pseudocode is as follows:

\begin{algorithm}[H]
\caption{KTD-CE Loss}
\begin{algorithmic}[1]
\REQUIRE Input predictions $\hat{y}$, ground truth labels $y$, scalar parameters $\alpha$ and $\beta$
\STATE Initialize $\text{CrossEntropyLoss}$ as $CELoss$
\STATE Initialize Kendall tau distance matrix $K \in R^{n \times n} $
\STATE Initialize offset matrix $ O\in R^{n \times n}$ with zeros
\STATE offset matrix requires gradient
\STATE $ce\_loss \gets CELoss(\hat{y}, y)$
\STATE Compute the argmax indices $\hat{y}_{\text{argmax}} \gets \text{argmax}(\hat{y}, \text{dim}=1)$
\STATE Retrieve Kendall Tau values: $kt\_values \gets (K + O)[\hat{y}_{\text{argmax}}, y]$
\STATE Compute $kt\_distance\_loss \gets \text{mean}(kt\_values)$
\STATE Compute L1 norm regularization: $l1\_norm \gets \sum | O |$
\RETURN $ce\_loss + \alpha \cdot kt\_distance\_loss + \beta \cdot l1\_norm$
\end{algorithmic}
\end{algorithm}

The loss function consists of three components:

\begin{itemize}
    \item \textbf{Cross Entropy Loss}
    
    The cross-entropy loss quantifies the difference between the predicted probability distribution and the true distribution in classification tasks. It is expressed as : 
    \begin{equation}
    \mathcal{L}_{\mathrm{CE}}=-\sum_{i=1}^{n}{y_ilog{\left(\widehat{y_i}\right)}}
    \end{equation}
    where $y_i$ is the true label, $\widehat{y_i}$ is the model's predicted probability, and $n$ denotes the number of classes.
    
    \item \textbf{Kendall Tau Distance Loss}
    
    The Kendall tau distance loss measures the difference between predicted and true orderings, calculated as:
   \begin{equation}
\begin{aligned}
\mathcal{L}_{\mathrm{KTD}} &= K\left[\arg\max(y_i), \arg\max(\widehat{y_i})\right] \\
&\quad + O\left[\arg\max(y_i), \arg\max(\widehat{y_i})\right]
\end{aligned}
\end{equation}
    where $K$ is a predefined, fixed Kendall Tau distance matrix , and $O$ is a trainable distance offset matrix of the same size as $K$. The notation $\left[i,j\right]$ denotes the corresponding coordinates in the matrix. While $K$ remains non-trainable, $O$ is adjustable to refine the Kendall tau distance.

    \item \textbf{Regularization Term}

    To prevent the distance offset matrix $O$ from becoming overly biased toward a specific ordering type, we introduce an L1 regularization term:
     \begin{equation}
    \mathcal{L}_{l_1}=\sum_{i=1}^{n}\sum_{j=1}^{n}\left|O\left[i=j\right]\right|
    \end{equation}
    
\end{itemize}

The complete form of the KTD-CE Loss is:
   \begin{equation}
  \mathcal{L}_{\mathrm{KTD-CE}}=\mathcal{L}_\mathrm{CE}+\alpha\mathcal{L}_{\mathrm{KTD}}+\beta\mathcal{L}_{l_1}
    \end{equation}

where $\alpha$ balances the weight of the Kendall Tau distance loss, and $\beta$ controls the influence of the regularization term. Based on experimental adjustments, we set $\alpha=1$ and $\beta= 0.1$.

\section{Cinematology Embedding for SSO} \label{sec:5}

\subsection{Overview} \label{sec:5.1}
In AI-assisted video editing, we aim for the generated videos to demonstrate a certain level of professionalism. However, from a model learning perspective, defining and quantifying "professionalism" is challenging due to the lack of standardized criteria. In contrast, shot sequencing in film production is meticulously designed by professional directors, serving as a highly specialized example of editing. Studies \cite{cite13, cite14} have revealed a significant correlation between film genres and shot selection, with different film types often following distinct patterns in shot language, such as shot size, angle, and motion.

Based on this observation, we introduced Cinematology Embedding, which incorporates movie metadata and shot types as prior knowledge to guide the shot sequence ordering task. In this study, film genre was chosen as the key metadata, while shot types were categorized into four dimensions: shot size, angle, motion, and type. Since manually labeling shot types is time-consuming and labor-intensive, we utilized an automated shot analysis method \cite{cite15} to label shot types efficiently. We then implemented a Video Transformer-based architecture \cite{cite57}, converting film genre and shot type labels into additional input tokens for the model. This approach enables the model to leverage narrative structure and visual style, providing richer contextual information for shot ordering and enhancing its performance in terms of professionalism.

\subsection{AVE-Meta} \label{sec:5.2}

The movie clips in the AVE dataset \cite{cite6}, sourced from the MovieClip channel on YouTube, include shot-level annotations but lack corresponding metadata. To address this, we developed the AVE-Meta dataset, supplementing it with movie metadata. Specifically, we identified the source of each AVE movie clip and extracted corresponding metadata from the IMDb database, including the movie's title, director, genre, release year, and other relevant information. Table \ref{tab:3} presents a sample from AVE-Meta, where the scene ID corresponds to the unique identifier of each scene in the AVE dataset. In this study, we primarily used the Genre tag as prior knowledge to guide the shot sequence ordering task. However, other metadata tags in the AVE-Meta dataset can also be employed for various film analysis tasks, such as director identification \cite{cite54}, content description \cite{cite46}, and video keyword generation \cite{cite55}.

\begin{table*}[h!]
\centering
\caption{Movie metadata for a scene in AVE-Meta}
\label{tab:3}
\resizebox{0.8\textwidth}{!}{
\begin{tabular}{|l|l|}
\hline
\textbf{Scene ID} & oCWVOBuvA \\ \hline
\textbf{IMDB ID} & tt0040872 \\ \hline
\textbf{Title} & They Live by Night \\ \hline
\textbf{Description} & 
An escaped convict injured during a robbery falls in love with the woman \\
& who nurses him back to health, but their relationship seems doomed \\
& from the beginning. \\ \hline
\textbf{Rating Count} & 9,090 \\ \hline
\textbf{Rating Value} & 7.4 \\ \hline
\textbf{Content Rating} & Passed \\ \hline
\textbf{Genre} & Crime, Film-Noir, Romance \\ \hline
\textbf{Release Date} & November 5, 1949 \\ \hline
\textbf{Keywords} & lovers on the lam, pet dog, singing, christmas tree, \\
& sweet potato pie \\ \hline
\textbf{Duration} & PT1H35M (95 minutes) \\ \hline
\textbf{Main Actors} & Cathy O'Donnell, Farley Granger, Howard Da Silva \\ \hline
\textbf{Director} & Nicholas Ray \\ \hline
\textbf{Creators} & Charles Schnee, Nicholas Ray, Edward Anderson \\ \hline
\end{tabular}
}
\end{table*}

\begin{figure*}[ht]
 \centering 
 \includegraphics[width=1.7\columnwidth]{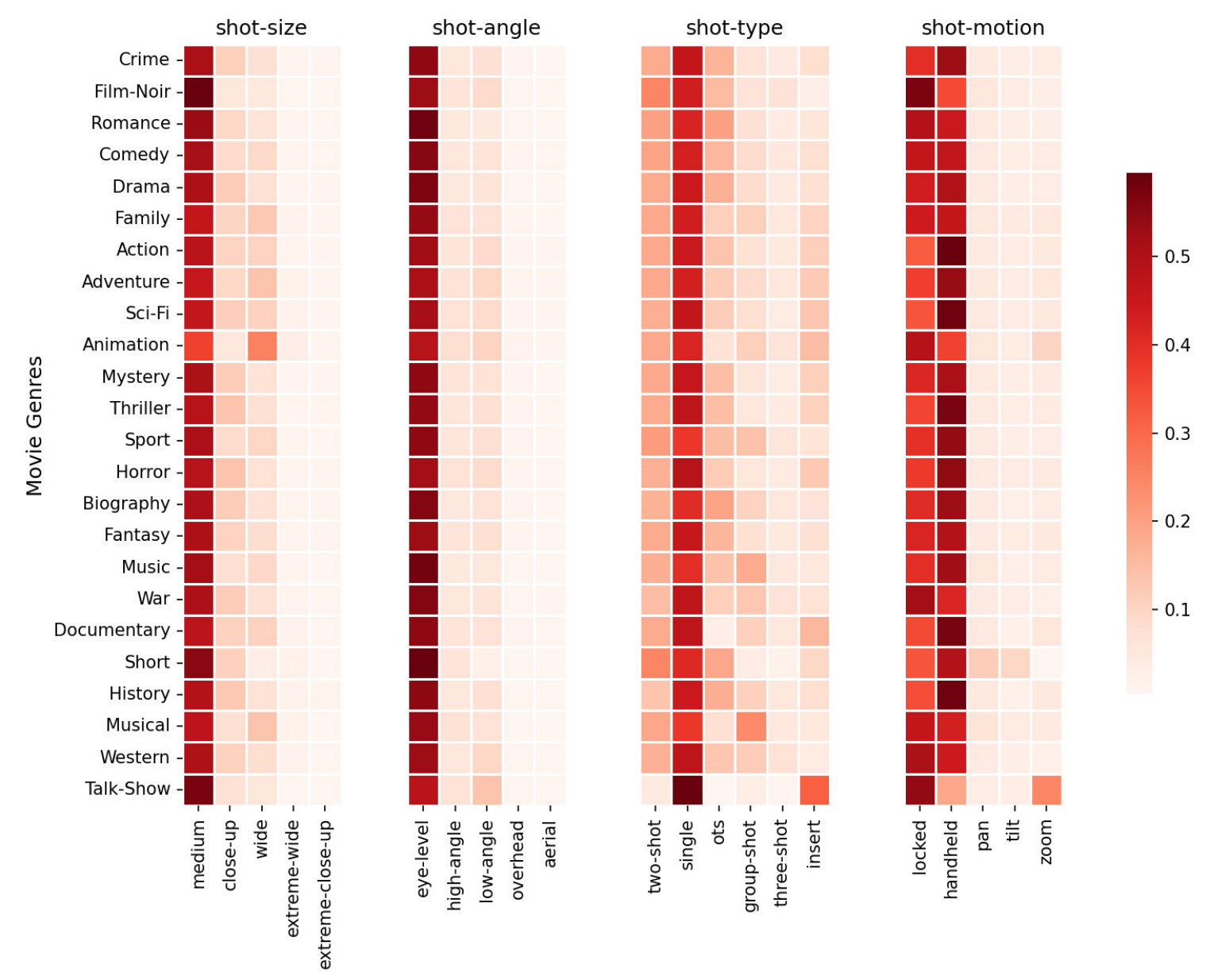}
 \caption{ Distribution of shot type frequencies across different film genres.}
 \label{fig:4}
\end{figure*}

Subsequently, we conducted an association analysis between the Genre tags in AVE-Meta and the shot type labels in AVE. Figure \ref{fig:4} illustrates the frequency of different shot types across various film genres (for a detailed description of shot types, refer to \cite{cite6}). This analysis confirmed the theories presented in \cite{cite13, cite14}. For instance, wide shots are significantly more common in animated films compared to other genres, while single shots are notably prevalent in talk shows.

\subsection{Shot Label Predict Methods} \label{sec:5.3}
\begin{figure*}[ht]
 \centering 
 \includegraphics[width=2\columnwidth]{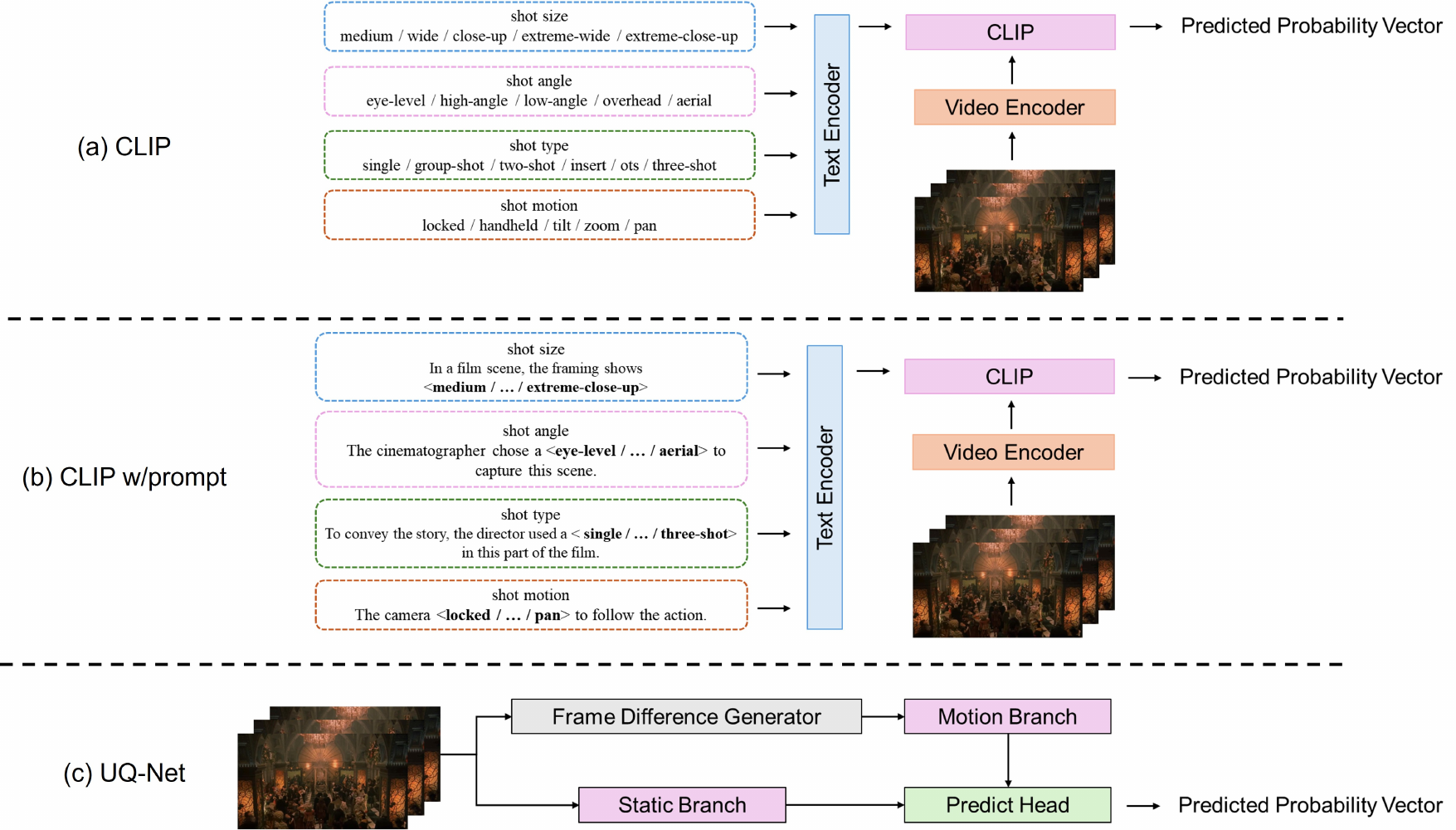}
 \caption{Shot Label Prediction Methods.(a) Directly using labels as text embeddings for CLIP (b) Using prompt-optimized labels as text embeddings for CLIP (c) Predicting shot labels using our proposed UQ-Net \cite{cite15}.}
 \label{fig:5}
\end{figure*}

Manual annotation of shot types is not only time-consuming and labor-intensive but also impractical for large-scale datasets. Drawing inspiration from \cite{cite58} and building on our previous research \cite{cite15}, we developed three automated methods for obtaining shot types.

Firstly, the CLIP model \cite{cite59} can project textual and visual information into a shared embedding space, enabling the generation of textual descriptions based on image content. As shown in Figure \ref{fig:5}(a), we used shot type labels (e.g., "wide shot," "close-up shot") as text embeddings and input them into the pre-trained CLIP model. By calculating the similarity between shots and text embeddings, we predicted the most likely shot type.

Secondly, the "CLIP with Prompt" method refines the CLIP input approach. As illustrated in Figure \ref{fig:5}(b), we designed prompts for each shot type and input them into the CLIP model along with the shot.

Lastly, leveraging our previous research \cite{cite15}, we employed the dual-stream network structure UQ-Net for shot analysis. As depicted in Figure \ref{fig:5}(c), this method processes each shot through both the motion branch and the static branch simultaneously. The motion branch captures dynamic features such as camera movement and motion trajectory, while the static branch focuses on static features, such as shot size and camera angle. The outputs from both branches are then fused to produce the final shot type prediction.

\begin{table}[h]
\centering
\caption{Shot Label Prediction Accuracy on the AVE Dataset (refer to \cite{cite15} for further details).}
\label{tab:4}
\begin{tabular}{ccccc}
\toprule
Method             & shot size     & shot angle    & shot motion   & shot type     \\
\midrule
Naive(V+A) \cite{cite6}      & 39.1          & 28.9          & 31.2          & 62.3          \\
Logitadj. (V+A) \cite{cite6} & 67.6          & 49.8          & 43.7          & 66.7          \\
R3D \cite{cite60}            & 67.5          & 85.4          & 70.2          & 55.2          \\
CLIP w/label \cite{cite59}   & 66.6          & 50.2          & 9.3           & 20.9          \\
CLIP w/prompt \cite{cite59}  & 41.0          & 28.6          & 2.6           & 24.4          \\
UQ-Net   \cite{cite15}        & \textbf{73.4} & \textbf{86.9} & \textbf{70.9} & \textbf{76.3}\\
\bottomrule
\end{tabular}
\end{table}

We subsequently analyzed the prediction performance of different methods across four shot label categories (shot size, shot angle, shot motion, and shot type), as shown in Table \ref{tab:4}. The results for Naive (V+A), Logitadj. (V+A), and R3D \cite{cite60} were directly referenced from \cite{cite6}. Our findings indicate that while the CLIP method yields relatively accurate predictions in certain cases, the specially trained UQ-Net consistently outperforms R3D in overall performance.

\subsection{Cinematology Embedding}\label{sec:5.4}

\begin{figure*}[ht]
 \centering 
 \includegraphics[width=2\columnwidth]{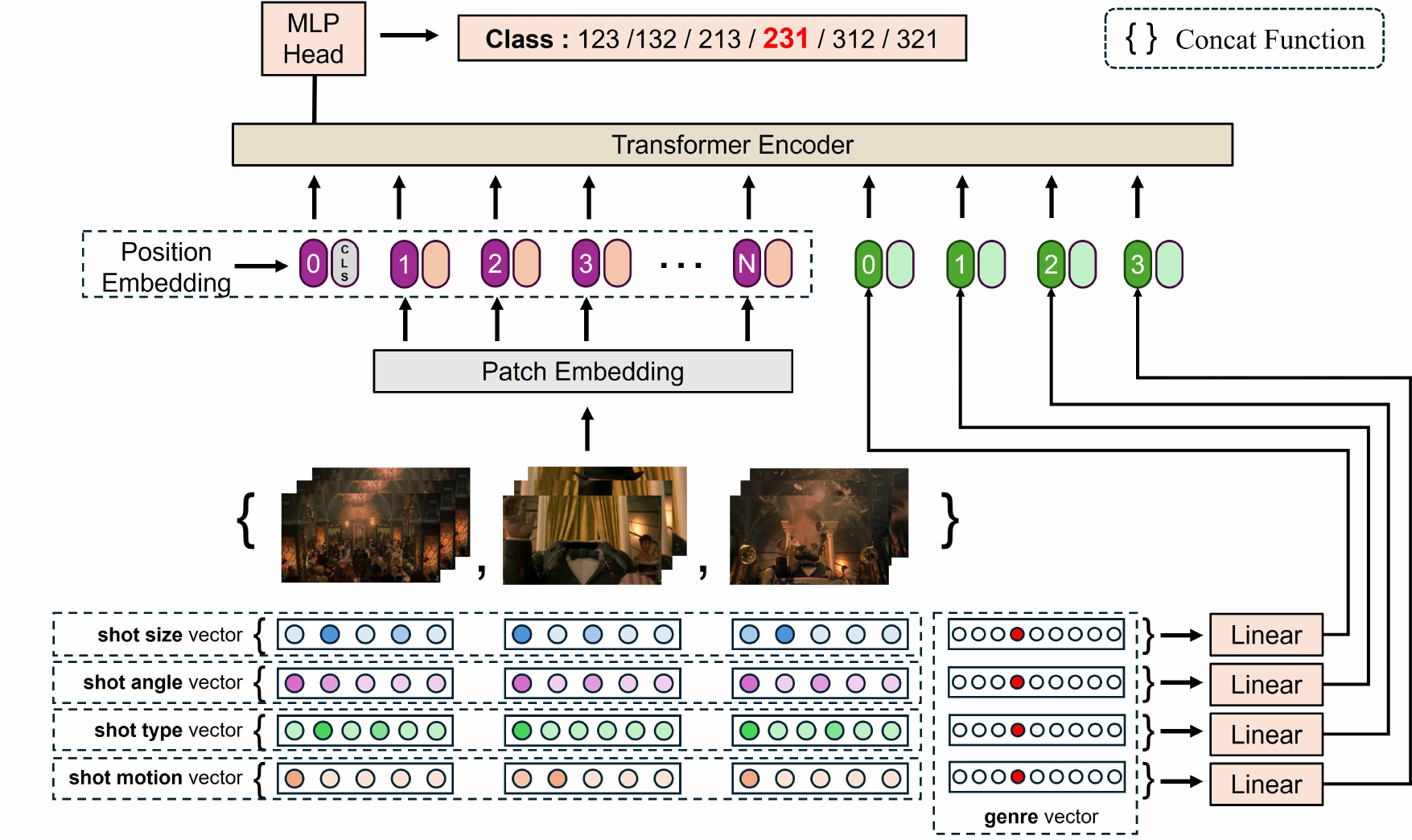}
 \caption{The overall architecture of the Cinematology Embedding. Shot label vectors of the same category are concatenated with movie genre labels. A linear layer then adjusts the concatenated vector's dimensions to align with the token dimensions obtained from the patch embedding. Finally, these fused tokens are input into the Video Transformer model for learning.}
 \label{fig:6}
\end{figure*}

To incorporate film genre labels and shot labels into the model for the SSO task, we adopted a Video Transformer-based architecture, as shown in Figure \ref{fig:6}, integrating shot grammar and film style as additional input tokens. We selected MViT-B \cite{cite57} as the base architecture, given its multi-scale capabilities to effectively capture different levels of visual features within the shots.

For each shot type, all shot label vectors in the sequence are first concatenated according to their types. Specifically, each shot’s label (shot size, shot angle, shot type, and shot motion) is transformed into a corresponding vector representation. The relevant one-hot film genre vector is then appended to each shot label vector, allowing each shot's features to reflect both its inherent attributes and the film's overall style.

Subsequently, these concatenated vectors are processed through a linear layer to match the token dimensions of the patch embedding. This step ensures that all tokens fed into the Transformer model—including visual tokens from original frames and the embedded shot grammar and film style tokens—maintain consistent feature dimensions, facilitating effective integration within the Transformer architecture. Throughout this process, if there are m different shot types, there will be m additional input tokens alongside the patch embedding tokens of the shot sequence.

By introducing Cinematology Embedding, the model can perform shot sequence ordering not only based on visual features but also by leveraging film genre and shot grammar information, enabling a more comprehensive and accurate judgment.

\section{Experiments}\label{sec:6}

\subsection{Implementation Detail}\label{sec:6.1}
We followed the TSN \cite{cite61} method during training, employing random uniform frame sampling for each shot. Each shot was divided into 8 uniform segments, from which one frame was randomly selected and resized to 224 × 224. As a result, each shot sequence sample comprised a total of 3 × 8 = 24 frames used as input to the model. During testing, each shot was similarly divided into 8 uniform segments, but the middle frame of each segment was fixed as the input. Details regarding the AVE-Order and ActivityNet-Order dataset splits are provided in Sections \ref{sec:3.2} and \ref{sec:3.3}.

Experiments were conducted using the PyTorch framework on an RTX 4090 GPU. The initial learning rate was set to 0.01, and we employed SGD as the optimizer, with a momentum of 0.5 and a weight decay coefficient of 0.0005. The model was trained over 30 epochs, with the learning rate reduced by a factor of 10 at the 15th and 25th epochs. For comparative purposes, we treated the shot sequence ordering as a general video classification task, utilizing various standard video classification models, with MViT \cite{cite57} serving as the baseline.

For the AVE-Order dataset, in addition to the three methods outlined in Section 5.3 for obtaining shot labels, we incorporated ground truth annotations from AVE \cite{cite6} to generate one-shot vectors as shot labels. In the ActivityNet-Order dataset, given that the data were sourced from user-uploaded videos with no clear shot type preferences, we assumed equal probability across all shot types without making any preference assumptions.

In Section \ref{sec:6.3}, models labeled \textbf{Ours} employed both the KTD-CE Loss and Cinematology Embedding, while others used the standard cross-entropy loss. An in-depth analysis of individual methods is presented in the ablation studies.

\subsection{Evaluation Metrics}\label{sec:6.2}

The following evaluation metrics were used to assess model performance in the SSO task:

\begin{itemize}
    \item \textbf{Top-1 Accuracy}

    This metric measures the proportion of instances where the model's predicted optimal ordering matches the true ordering, reflecting the model's overall accuracy in making the single best prediction. It is calculated as follows:
    \begin{equation}
    Top-1 \ Accuracy=\frac{1}{N}\sum_{i=1}^{N}\mathbb{1}(\widehat{y_i},y_i)
    \end{equation}
    Here, $N$ is the total number of samples, $\widehat{y_i}$ represents the model's predicted ordering, $y_i$ is the true ordering, and $\mathbb{1}(\cdot)$ is an indicator function that equals 1 when $\widehat{y_i}=y_i$ otherwise 0.

    \item \textbf{Top-k Accuracy}

    This metric evaluates the likelihood that the correct ordering is included within the top-k predictions of the model. In this study, $k=3$. The formula is:
    \begin{equation}
 Top-k \  Accuracy=\frac{1}{N}\sum_{i=1}^{N}\mathbb{1}(y_i\in \{ \widehat{y}_i^{(1)},\widehat{y}_i^{(2)},...,\widehat{y}_i^{(k)} \})
    \end{equation}
    Where $y_i\in \{ \widehat{y}_i^{(1)},\widehat{y}_i^{(2)},...,\widehat{y}_i^{(k)}\}$ denotes the top-$k$ predictions for the $i$-th sample.
    
    \item \textbf{Recall}

    This metric measures the proportion of true positive ordering labels correctly identified by the model, indicating its sensitivity to positive cases. It is calculated as:
    \begin{equation}
    Recall=\frac{TP}{TP+FN}
    \end{equation}
    where $TP$ represents true positives correctly identified, and $FN$ represents false negatives incorrectly classified as negative.

    \item \textbf{Precision}

   This metric measures the proportion of predicted positive ordering labels that are actual true positives, which is crucial when prioritizing prediction accuracy over quantity. It is calculated as:
    \begin{equation}
   Precision=\frac{TP}{TP+FP}
    \end{equation}
    Where $TP$ denotes true positives correctly identified, and $FP$ denotes false positives incorrectly classified as positive. 

    \item \textbf{Kendall Tau Distance}

    This metric evaluates the difference between the model's predicted ordering and the true ordering by counting the number of inverted pairs, thus reflecting ranking differences. A lower Kendall Tau Distance indicates a predicted ordering closer to the true ordering. The detailed calculation method is provided in Section \ref{sec:4.1}.
    
\end{itemize}

\subsection{Overall Results}\label{sec:6.3}

\subsubsection{AVE-Order} \label{sec:6.3.1}

\begin{table*}[ht]
\centering
\caption{Overall results on AVE-Order (↑indicates that a higher value is better, ↓ indicates that a lower value is better)}
\label{tab:5}
\resizebox{0.9\textwidth}{!}{
\begin{tabular}{llccccc}
\toprule
Method                & Note                                                             & \multicolumn{1}{c}{\begin{tabular}[c]{@{}c@{}}Top-1 \\ Accuracy↑\end{tabular}} & \multicolumn{1}{c}{\begin{tabular}[c]{@{}c@{}}Top-3 \\ Accuracy↑\end{tabular}} & \multicolumn{1}{c}{\begin{tabular}[c]{@{}c@{}}Kendall tau   \\ distance↓\end{tabular}} & \multicolumn{1}{c}{Recall↑} & \multicolumn{1}{c}{Precision↑} \\
\midrule
Random                &                                                                  & 16.60                                                                          & 50.00                                                                          & 1.500                                                                                  & 16.66                       & 16.66                          \\
\midrule
Human                 & report in \cite{cite6}                                                    & 39.9                                                                           & ——                                                                             & ——                                                                                     & ——                          & ——                             \\ 

R3D early fusion \cite{cite60}  & report in \cite{cite6}                                                    & 25.7                                                                           & ——                                                                             & ——                                                                                     & ——                          & ——                             \\
R3D late fusion \cite{cite60}   & report in \cite{cite6}                                                    & 21.5                                                                           & ——                                                                             & ——                                                                                     & ——                          & ——                             \\
\midrule
SlowFast \cite{cite62}          &                                                                  & 25.29                                                                          & 62.03                                                                          & 1.474                                                                                  & 25.38                       & 24.42                          \\
R3D \cite{cite60}               &                                                                  & 25.44                                                                          & 62.13                                                                          & 1.469                                                                                  & 25.33                       & 26.33                          \\
\midrule
MViT \cite{cite57}             & \begin{tabular}[c]{@{}l@{}}pre-trained with \\ K400\end{tabular} & 32.04                                                                          & 66.82                                                                          & 1.345                                                                                  & 32.04                       & 37.03                          \\
VideoSwin \cite{cite63}         & \begin{tabular}[c]{@{}l@{}}pre-trained with \\ K400\end{tabular} & 27.21                                                                          & 64.36                                                                          & 1.474                                                                                  & 27.22                       & 28.65                          \\
VideoMAE \cite{cite64}          & \begin{tabular}[c]{@{}l@{}}pre-trained with \\ K400\end{tabular} & 25.29                                                                          & 62.03                                                                          & 1.474                                                                                  & 25.37                       & 24.42                          \\
\midrule
Ours / ground truth   &                                                                  & 34.65                                                                          & 68.79                                                                          & 1.267                                                                                  & 34.65                       & 35.47                          \\
Ours / uqn-predict    &                                                                  & 35.28                                                                          & 69.81                                                                          & 1.251                                                                                  & 35.27                       & 36.54                          \\
Ours / clip-predict   &                                                                  & 34.62                                                                          & 68.52                                                                          & 1.278                                                                                  & 34.56                       & 35.59                          \\
Ours / clip/p-predict &                                                                  & 33.30                                                                          & 68.00                                                                          & 1.325                                                                                  & 33.25                       & 34.59          \\
\bottomrule
\end{tabular}
}
\end{table*}

Table \ref{tab:5} shows the experimental results of various methods on the AVE-Order dataset. \textbf{Random} indicates the model's performance without training, while \textbf{Human} refers to the human-performed experimental results from \cite{cite6}. Both \textbf{R3D early fusion} and \textbf{R3D late fusion} results are also sourced from \cite{cite6}. As the original paper did not provide training code, we retrained R3D with early fusion on the AVE-Order dataset, and the results closely matched the performance reported in \cite{cite6}, making it equivalent to R3D early fusion. Therefore, the human results can serve as a comparative benchmark.

In the experiments, the traditional 2D convolutional network \textbf{SlowFast} \cite{cite62} and the 3D convolutional network \textbf{R3D} \cite{cite60} exhibited suboptimal performance. Similarly, \textbf{VideoMAE} \cite{cite64}, despite showing excellent performance in tasks like action recognition \cite{cite64}, did not achieve the anticipated results in the SSO task of this study.

Among Video Transformer-based methods, we evaluated \textbf{MViT} \cite{cite57} and \textbf{VideoSwin} \cite{cite63}. \textbf{MViT} achieved a 7\% improvement in Top-1 Accuracy over previous methods and was therefore chosen as the base architecture for Cinematology Embedding. However, \textbf{MViT} did not demonstrate a significant advantage in the Kendall Tau Distance metric we proposed. For the shot sequence ordering task, we argue that, compared to solely focusing on Top-1 Accuracy, Kendall Tau Distance offers a more comprehensive reflection of the model's learning performance.

Regarding our method, the \textbf{Ours / uqn-predict} model used shot label vectors predicted by UQ-Net \cite{cite15}, yielding exceptional performance in both Top-1 Accuracy and Kendall Tau Distance, even surpassing \textbf{Ours / ground truth}, which utilized hard labels. However, it still fell short of achieving human-level results. Furthermore, it is evident that the accuracy of shot labels critically influences the overall performance of the model, even when other conditions remain constant.

\subsubsection{ActivityNet-Order}\label{sec:6.3.2}

\begin{table*}[ht]
\centering
\caption{Overall results on ActivityNet-Order}
\label{tab:6}
\resizebox{0.9\textwidth}{!}{
\begin{tabular}{llccccc}
\toprule
Method                & Note                                                               & \multicolumn{1}{c}{\begin{tabular}[c]{@{}c@{}}Top-1 \\ Accuracy↑\end{tabular}} & \multicolumn{1}{c}{\begin{tabular}[c]{@{}c@{}}Top-3 \\ Accuracy↑\end{tabular}} & \multicolumn{1}{c}{\begin{tabular}[c]{@{}c@{}}Kendall tau   \\ distance↓\end{tabular}} & \multicolumn{1}{c}{Recall↑} & \multicolumn{1}{c}{Precision↑} \\
\midrule
Random                &                                                                    & 16.60                                                                          & 50.00                                                                          & 1.500                                                                                  & 16.60                       & 16.60                          \\
\midrule
SlowFast \cite{cite62}          &                                                                    & 23.92                                                                          & 60.99                                                                          & 1.484                                                                                  & 23.75                       & 26.75                          \\
R3D \cite{cite60}               &                                                                    & 24.40                                                                          & 60.62                                                                          & 1.484                                                                                  & 24.23                       & 20.66                          \\
\midrule
MViT \cite{cite57}              & \begin{tabular}[c]{@{}l@{}}pre-trained with   \\ K400\end{tabular} & 45.56                                                                          & 76.20                                                                          & 0.974                                                                                  & 45.55                       & 48.81                          \\
VideoSwin \cite{cite63}         & \begin{tabular}[c]{@{}l@{}}pre-trained with   \\ K400\end{tabular} & 31.55                                                                          & 66.41                                                                          & 1.343                                                                                  & 31.54                       & 33.12                          \\
VideoMAE \cite{cite64}          & \begin{tabular}[c]{@{}l@{}}pre-trained with   \\ K400\end{tabular} & 34.11                                                                          & 68.41                                                                          & 1.297                                                                                  & 34.15                       & 40.36                          \\
\midrule
Ours / uqn-predict    &                                                                    & 51.41                                                                          & 78.76                                                                          & 0.862                                                                                  & 51.43                       & 51.68                          \\
Ours / clip-predict   &                                                                    & 51.17                                                                          & 79.13                                                                          & 0.875                                                                                  & 51.14                       & 51.91                          \\
Ours / clip/p-predict &                                                                    & 51.02                                                                          & 78.99                                                                          & 0.882                                                                                  & 51.08                       & 52.96                   \\
\bottomrule
\end{tabular}
}
\end{table*}

Table \ref{tab:5} presents the experimental results for various methods on the ActivityNet-Order dataset. The traditional 2D convolutional network \textbf{SlowFast} \cite{cite62} and the 3D convolutional network \textbf{R3D} \cite{cite60} delivered poor performance. \textbf{MViT} \cite{cite57} achieved improvements of 21.1 (Top-1 Accuracy↑) / 0.51 (Kendall Tau Distance↓) compared to \textbf{R3D} and further improved by 14.0 / 0.37 and 11.4 / 0.32 relative to \textbf{VideoSwin} and \textbf{VideoMAE}, respectively. Our method further enhanced \textbf{MViT}'s performance by 5.8 / 0.11.

Notably, even in the high-quality user-generated ActivityNet-Order dataset, incorporating film grammar as prior knowledge proved effective. This finding suggests that many video editing techniques have roots in film production, indicating that film production experience and grammar offer valuable guidance for editing non-professional video content.

\subsection{Ablation Studies}\label{sec:6.4}
To assess the effectiveness of the proposed KTD-CE Loss and Cinematology Embedding, we conducted ablation experiments comparing the model's performance on the AVE-Order and ActivityNet-Order datasets using different loss functions and embedding strategies. This analysis aimed to determine the specific contributions of these improvements to SSO task.

\subsubsection{Effective of KTD-CE Loss}\label{sec:6.4.1}

\begin{table*}[]
\caption{The effect of different loss functions on the SSO task.}
\label{tab:7}
\centering
\resizebox{0.9\textwidth}{!}{
\begin{tabular}{lccc|ccc}
\toprule
Method      & \multicolumn{3}{c|}{AVE-Order}                                                                                                                                                             & \multicolumn{3}{c}{ActivityNet-Order}                                                                                                                                                      \\
            & \begin{tabular}[c]{@{}c@{}}Top-1 \\ Accuracy↑\end{tabular} & \begin{tabular}[c]{@{}c@{}}Top-3 \\ Accuracy↑\end{tabular} & \begin{tabular}[c]{@{}c@{}}Kendall tau \\ distance↓\end{tabular} & \begin{tabular}[c]{@{}c@{}}Top-1 \\ Accuracy↑\end{tabular} & \begin{tabular}[c]{@{}c@{}}Top-3 \\ Accuracy↑\end{tabular} & \begin{tabular}[c]{@{}c@{}}Kendall tau \\ distance↓\end{tabular} \\ \midrule
KTD Loss    & 17.17                                                      & 50.08                                                      & 1.481                                                            & 17.21                                                      & 50.36                                                      & 1.481                                                            \\
CE Loss     & 34.56                                                      & 68.77                                                      & 1.284                                                            & 51.78                                                      & 79.16                                                      & 0.865                                                            \\
KTD-CE Loss & 35.28                                                      & 69.81                                                      & 1.251                                                            & 51.41                                                      & 78.76                                                      & 0.862   \\ \bottomrule                                                        
\end{tabular}
}
\end{table*}

Table \ref{tab:7} shows the model's performance on the two datasets with varying loss functions, while maintaining the Cinematology Embedding. The results indicate that using only the KTD Loss led to poor performance on both datasets, suggesting that it is insufficient to fully capture shot sequence ordering relationships on its own. On the AVE-Order dataset, the KTD-CE Loss outperformed the CE Loss across all metrics. In contrast, on the ActivityNet-Order dataset, although the KTD-CE Loss exhibited slightly lower Top-1 Accuracy compared to the CE Loss, it performed better on the Kendall Tau Distance metric. These findings suggest that the KTD-CE Loss effectively combines the strengths of both KTD Loss and CE Loss, retaining the benefits of classification loss while optimizing sequence ordering relationships, thereby delivering a more comprehensive performance improvement in the SSO task.

\subsubsection{Effective of Cinematology Embedding}\label{sec:6.4.2}

\begin{table*}[]
\caption{The effect of different Cinematology Embedding inputs on the SSO task.}
\label{tab:8}
\centering
\resizebox{1\textwidth}{!}{
\begin{tabular}{lccc|ccc}
\toprule
Method                     & \multicolumn{3}{c|}{AVE-Order}                                                                                                                                                                                                                          & \multicolumn{3}{c}{ActivityNet-Order}                                                                                                                                                                                                                  \\
                           & \multicolumn{1}{c}{\begin{tabular}[c]{@{}c@{}}Top-1 \\ Accuracy↑\end{tabular}} & \multicolumn{1}{c}{\begin{tabular}[c]{@{}c@{}}Top-3 \\ Accuracy↑\end{tabular}} & \multicolumn{1}{c|}{\begin{tabular}[c]{@{}c@{}}Kendall tau \\ distance↓\end{tabular}} & \multicolumn{1}{c}{\begin{tabular}[c]{@{}c@{}}Top-1 \\ Accuracy↑\end{tabular}} & \multicolumn{1}{c}{\begin{tabular}[c]{@{}c@{}}Top-3 \\ Accuracy↑\end{tabular}} & \multicolumn{1}{c}{\begin{tabular}[c]{@{}c@{}}Kendall tau \\ distance↓\end{tabular}} \\ \midrule
w/ Cinematology Embedding  & 35.28                                                                          & 69.81                                                                          & 1.251                                                                                 & 51.41                                                                          & 78.76                                                                          & 0.862                                                                                \\
w/o Cinematology Embedding & 32.31                                                                          & 66.38                                                                          & 1.316                                                                                 & 50.80                                                                          & 78.63                                                                          & 0.875                                                                                \\
w/o genre label            & 34.97                                                                          & 69.48                                                                          & 1.258                                                                                 & 51.13                                                                          & 78.77                                                                          & 0.870                                                                                \\
w/o shot labels            & 33.41                                                                          & 66.86                                                                          & 1.303                                                                                 & 50.45                                                                          & 78.48                                                                          & 0.877                                 \\
\bottomrule
\end{tabular}
}
\end{table*}

Table \ref{tab:8} presents the model's performance on the two datasets when adjusting the Cinematology Embedding inputs, while keeping the KTD-CE Loss constant. To evaluate the impact of the Cinematology Embedding, we examined model performance under three scenarios: without Cinematology Embedding, excluding the genre label, and excluding shot labels.

The experimental results demonstrate that using the complete Cinematology Embedding significantly enhanced model performance. On both the AVE-Order and ActivityNet-Order datasets, the model achieved optimal performance in both Top-1 Accuracy and Kendall Tau Distance metrics. Removing the genre label caused a slight decline in performance, though the impact was minimal. However, the exclusion of shot labels resulted in a more substantial decrease in performance. This finding indicates that shot category information plays a more critical role than the genre label within Cinematology Embedding and can be considered a form of general cinematic grammar as prior knowledge. Overall, the Cinematology Embedding proved to be effective in enhancing the model's performance for the SSO task.
\section{Challenge} \label{sec:7}

This study provides a systematic review of the SSO task, however, it represents only a preliminary exploration, and several issues remain unresolved.

Firstly, the current research primarily focuses on the 3-shot sequence ordering task, but in practical video editing, sequences are often longer, and shot order combinations are more complex. As the number of shots increases, the permutations of sequences grow factorially, significantly escalating the task's difficulty and computational complexity. Therefore, extending to multi-shot sequence ordering and managing longer, more intricate shot sequences is a critical issue that future research must address.

Secondly, the interpretability of shot sequence ordering requires further enhancement. Although the proposed KTD-CE Loss improves the model's ordering capability to some extent, existing interpretability methods like Grad-CAM \cite{cite65} may not be suitable for the SSO task due to its focus on global temporal relationships, whereas methods such as Grad-CAM typically concentrate on single frames or local regions.

Lastly, in real-world applications of AI-assisted video editing, shot sequence ordering often functions as a zero-shot task, where editors must handle unfamiliar scenes or video content. In such cases, the model must be capable of performing shot sequence ordering based on existing knowledge, even without training samples. This requirement places higher demands on the model's generalization ability, knowledge transfer, and understanding of visual language.

\section{Conclusion}\label{sec:8}
 
This study systematically explored the Shot Sequence Ordering task for the first time. To address the current data scarcity, we curated and developed two publicly accessible benchmark datasets—AVE-Order and ActivityNet-Order—thereby filling a critical data gap in this field. Additionally, we introduced Kendall Tau Distance as an evaluation metric for the SSO task and proposed the KTD-CE loss function tailored for SSO tasks. To further improve shot ordering accuracy, we also introduced the Cinematology Embedding method, which integrates movie metadata and shot labels as prior knowledge into the SSO model.

\bibliographystyle{IEEEtran}

\end{document}